# The Need for and Feasibility of Alternative Ground Robots to Traverse Sandy and Rocky Extraterrestrial Terrain

*Chen Li* and Kevin Lewis*

**Robotic spacecrafts have helped expand the reach for many planetary exploration missions. Most ground mobile planetary exploration robots use wheeled or modified wheeled platforms. Although extraordinarily successful at completing intended mission goals, because of the limitations of wheeled locomotion, they have been largely limited to benign, solid terrain and avoided extreme terrain with loose soil/sand and large rocks. Unfortunately, such challenging terrain is often scientifically interesting for planetary geology. Although many animals traverse such terrain at ease, robots have not matched their performance and robustness. This is in major part due to a lack of fundamental understanding of how effective locomotion can be generated from controlled interaction with complex terrain on the same level of flight aerodynamics and underwater vehicle hydrodynamics. Early fundamental understanding of legged and limbless locomotor–ground interaction has already enabled stable and efficient bioinspired robot locomotion on relatively flat ground with small obstacles. Recent progress in the new field of terradynamics of locomotor–terrain interaction begins to reveal the principles of bioinspired locomotion on loose soil/sand and over large obstacles. Multilegged and limbless platforms using terradynamics insights hold the promise for serving as robust alternative platforms for traversing extreme extraterrestrial terrain and expanding the reach in planetary exploration.**

Robotic spacecraft have vastly increased our ability to explore extraterrestrial surfaces.[1–6] Mobile robots have enabled exploration beyond a static landing site and allowed discovery-driven investigations on both the Moon and Mars. They have helped us understand the geologic history and surface environments of both bodies, conducting scientific campaigns analogous in many ways to that of a terrestrial field geologist.

Since the first deployments on the Moon nearly half a century ago, mobile planetary exploration robots have progressively increased in their capabilities and enabled us to access a range of scientific targets on extraterrestrial surfaces.[2–5] To date, all of the successfully landed lunar and Martian mobile exploration robots, as well as the majority of those being developed, have adopted a conventional wheeled rover platform with a variety of architectures.[5,7] This is not surprising because one of the highest priority considerations for space applications is to maximize mechanical reliability, where wheeled platforms have excelled.[2,3,7,8] These missions have been hugely successful, often exceeding mission design lifetime and traverse distance. The 6-wheel Mars Exploration Rover Opportunity currently holds the planetary rover distance record, driving over 45 km across Meridiani Planum.[9] Similarly, the Soviet Union's 8-wheel Lunokhod 2 rover traversed 39 km across the surface of the Moon in 1973.[10] The recent 6-wheel Chinese Yutu and Yutu-2 lunar rovers have travelled hundreds of meters on the surface of the Moon.[11]

Although these rovers have had an impressive track record exploring both the Moon and Mars, their missions have revealed significant limitations faced by wheel-based mobility systems, which hinder scientific exploration. For example, the Spirit Mars Exploration Rover ultimately reached the end of its mission due to a low power state after becoming embedded in a patch of loose soil at a location known as "Troy." The ferric sulfate dominated soil at this site had very low cohesion, thus being mechanically weak, and extended to a depth comparable to the wheel radius.[12] Unfortunately, this deposit was hidden beneath a weakly indurated soil crust, rendering the hazard hidden until the rover was already embedded.[9] The challenge of extricating Spirit was made more difficult due to the failure of one of its six wheels earlier in the mission, requiring modified driving strategies.[12] The Opportunity rover had similar challenges navigating ubiquitous large aeolian ripples in Meridiani Planum. In particular, it was stuck for an extended time embedded in the loose sand of the "Purgatory" ripple[13] (**Figure 1**A).

More recently, the Curiosity Mars rover has incurred significant wheel damage along its traverse due to angular rocks protruding from the surface, which punctured the thin aluminum

C. Li
Department of Mechanical Engineering
Johns Hopkins University
3400 N. Charles St, Baltimore 20218, MD, USA
E-mail:

K. Lewis
Department of Earth & Planetary Sciences
Johns Hopkins University
3400 N. Charles St, Baltimore 20218, MD, USA

The ORCID identification number(s) for the author(s) of this article can be found under https://doi.org/10.1002/aisy.202100195.

skin of the wheels (as can be seen in Figure 1B).[14] For Curiosity, two strategies were developed to minimize wheel damage. First, new wheel commanding procedures were developed to respond to the varying terrain underneath the rover, dubbed Traction Control. Previously, wheels were commanded at a fixed rotation rate regardless of varying topography. This new traction control strategy rotates the wheels to drive at different speeds when one or more wheels become jammed, which reduces opposing forces that damage the wheels.[15] Second, the rover traverse route was altered and limited to favorable terrain types with minimal wheel hazards.[16] Combined, these two strategies have reduced subsequent damage to Curiosity's wheels. Curiosity also had its wheels sunk deep in soft sand at a location dubbed "Hidden Valley" (Figure 1B). Mission planners had hoped to drive across the rippled sand of Hidden Valley to protect the rover's wheels but decided to back out and stick to harder ground.[17] These competing challenges of sinking into loose soil or incurring wheel damages on harder ground have limited accessible terrain types.

Unfortunately, reaching many of the most scientifically interesting sites requires traversing hazardous terrain types. Eroding cliff faces can offer freshly exposed outcrop, but this erosion often leads to accumulated angular rock fragments at the base of the cliff. Some of these rocks are highly cluttered (Figure 1C). Some are much larger than the rovers (Figure 1D).[18] Some are exposed on steep slopes at or near the angle of repose ($\approx 35°$) at which loose sediment is stable.[18] On Mars in particular, such steep topographic features often also accumulate loose, windblown sand deposits (Figure 1D,E).

Orbital reconnaissance is now able to characterize terrain down to submeter scales for the Moon and Mars and guide planning of robot exploration routes through modest terrain.[14] Terrain traversability is taken into account even prior to choosing a rover landing site, as was done for the Mars 2020 Perseverance rover.[19] The landing site selection process for this and previous rovers was community-led beginning with a large number of proposed sites and proceeding through several stages of down-selection, ultimately resulting in the selection of Jezero crater as the rover landing site. Although science priorities were the initial and primary drivers in the selection process, traversability from the landing site to the regions of scientific interest was taken into account in later stages of the down-selection process.[20] For example, the proposed but not ultimately selected landing site in Holden crater was noted for its drive difficulty due to abundant aeolian ripples that would have to be crossed along the traverse route.[20] The limitation of accessible terrain types for wheeled rovers has unfortunately precluded scientifically interesting targets in a number of cases.

Ultimately, to access a wider range of scientific targets, we need alternative robots to complement wheeled rovers with the ability to traverse a broader variety of terrain types and obstacles on the surface while minimizing the risks of being immobilized or significantly damaged. This need is even more crucial for future robots that will explore planetary bodies that are less well characterized from orbit, such as Venus, Titan, or Europa.

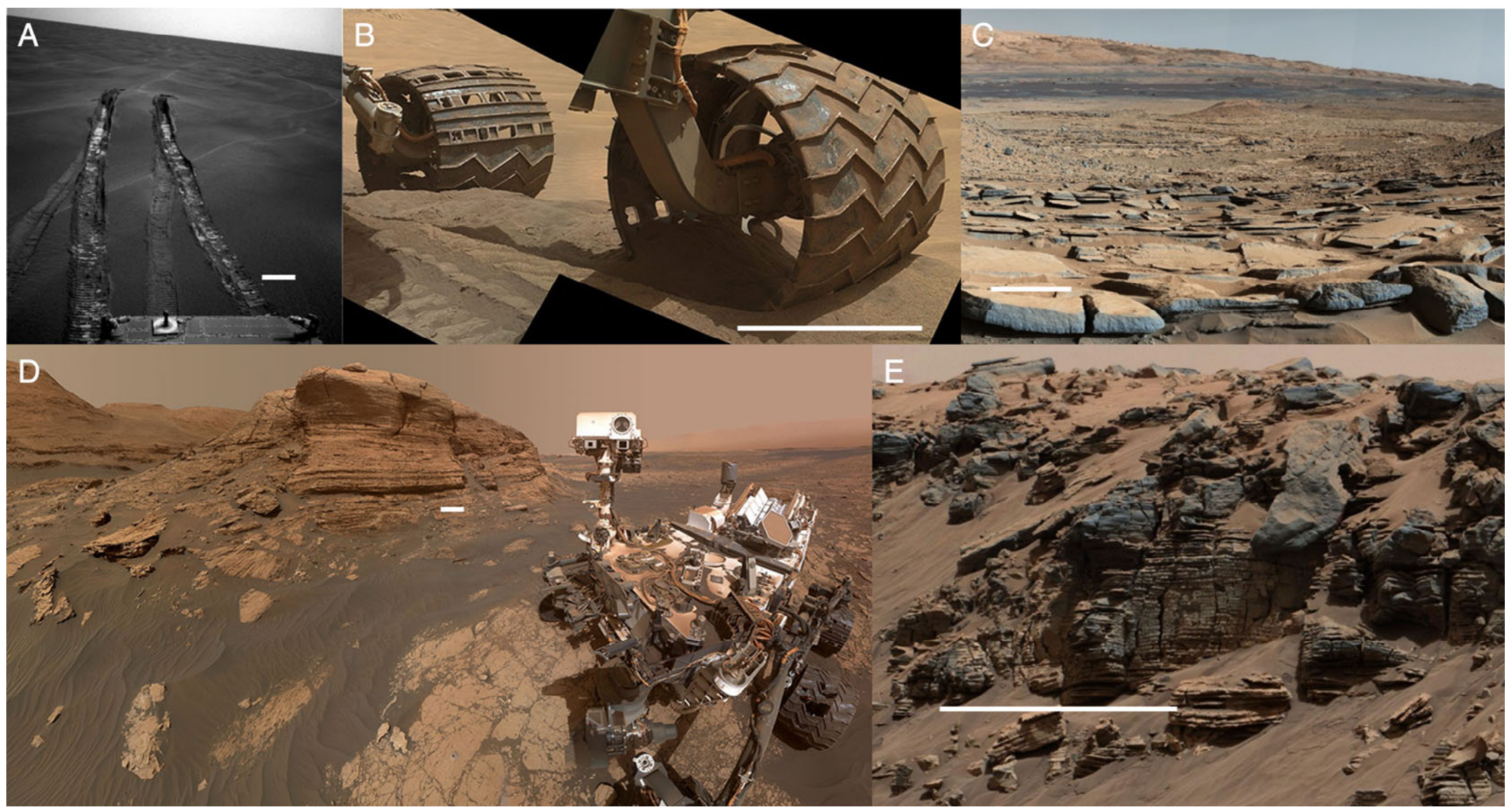

**Figure 1.** Scientifically interesting yet challenging extraterrestrial terrain for wheeled rovers to traverse. Scale bars in all panels are 0.5 m (rover wheel diameter). A) Opportunity rover at "Purgatory ripple," sol 446. The rover was embedded in this 30 cm high aeolian ripple for several weeks, due to the low cohesion sand, and subsequently avoided comparably sized ripples along the traverse (photojournal.jpl.nasa.gov/catalog/PIA07999). B) Curiosity with its wheels deep in soft sand at Hidden Valley, sol 711 (https://mars.nasa.gov/raw_images/188614/?site=msl, https://mars.nasa.gov/raw_images/188616/?site=msl). C) Curiosity Mastcam mosaic at the "Square Top" outcrop in the Kimberley region, on sol 580. The sandstone unit in this region is scientifically interesting, but inaccessible to the rover due to puncture hazards to its wheels from sharp angular rocks. Image shown is false color to accentuate color differences within the scene. For scale, the block in the lower left is roughly 1 m across (mars.nasa.gov/resources/7505/strata-at-base-of-mount-sharp). D) Curiosity Rover at the 4 m-high outcrop "Mont Mercou," sol 3060. Note the abundant sand cover and jagged rocks at the base of the cliff, along with Curiosity's wheel damage incurred to date over the course of the mission (https://photojournal.jpl.nasa.gov/catalog/PIA24543). E) Mosaic of the wall of "Hidden Valley" as imaged by the Curiosity rover on sol 712. Accumulated large angular rocks and loose windblown sand deposits make it difficult rock outcrops of scientific interest, such as the one shown in the center of the frame (photojournal.jpl.nasa.gov/catalog/PIA19074). Image credits: NASA/JPL-Caltech/MSSS.

Most robots, including planetary rovers, have approached locomotion in complex environments as a problem of obstacle avoidance. For example, a Mars rover uses orbital and onboard vision to create a geometric map of the environment (**Figure 2**A), uses computer vision to classify and identify terrain that is likely to be challenging due to obstacles (Figure 2B), and then plans and controls itself to follow a safe path to move around obstacles by transitioning between driving modes (Figure 2C).[21] This approach requires that obstacles are sparse and locomotor–ground interaction is well understood and readily controlled. It is because we understand tire dynamics[22] for rigid ground and terramechanics for deformable ground[8,23,24] as well as we do that wheeled vehicles and robots move so well on paved roads and off-road terrain. However, loose sand/soil can result in sinkage comparable to the diameter of small rover wheels (Figure 1A,B), whose interaction with deformable ground is not well described by classical terramechanics,[25–28] and rocks can be simply too cluttered to avoid (e.g., Figure 1C) or too steep to climb for wheels (Figure 1D,E). Thus, they are identified as obstacles to be avoided, even though there lie some of the most desired scientific targets. Although alternative platforms such biologically inspired robots are potentially suitable for traversing them, we still lack a fundamental understanding of locomotor–terrain interaction for such terrain on the same level of tire dynamics and terramechanics, based on which controlled locomotion can be generated by transitioning between desired locomotor modes (e.g., **Figure 3**A).

Recent research has alleviated these limitations to some extent. For loose soil/sand, refined terramechanical models[25–28] and new methods[29,30] based on granular resistive force theory[31,32] better predict sinkage and terrain traversability for small rover wheels. Some new rover platforms added leg-like degrees of freedom controlling wheels to generate various "gaits" (sequences of leg and wheel actuation and coordination), which can be tuned to slowly excavate itself after digging into loose sand.[33] For rocky terrain, many wheeled rovers have been developed to adopt a "wheel-on-leg" design, with actuated leg-like active suspension systems that can lift wheels onto obstacles of varying height while keeping the chassis upright.[7,34–41] Given these progresses, wheeled platforms are still inherently less suited for highly challenging terrain of scientific interest.

To enable access to a broader range of terrain, besides flying[42] and underwater[43,44] robots, many nonwheeled ground robot platforms have been developed, most of which specializes in a single mode of locomotion[45–59] (for a review, see Thoesen and Marvi[7]). These primarily include: tracked,[52] multilegged,[48,49,54,56] humanoid biped,[53,57] screw-propelled,[55,58] snake-like,[55] tensegrity-based robots,[59] scaling,[50] jumping,[45,47] and rappelling[1,6,46,51] robots. A recent conference on Planetary Exploration Robots: Challenges and Opportunities has featured several unconventional robotic platforms.[60] As the cost of space access reduces,[61] opportunities will increase for the testing and eventual deployment of these alternative platforms that complement rovers to enable access to a broader range of terrains for scientific exploration on extraterrestrial bodies. However, a major challenge remains for a nonwheeled platform to become capable of multimodal locomotion to traverse a diversity of terrain. There is a lack of principled understanding of how to compose transitions across different locomotor modes analogous to those for wheels.

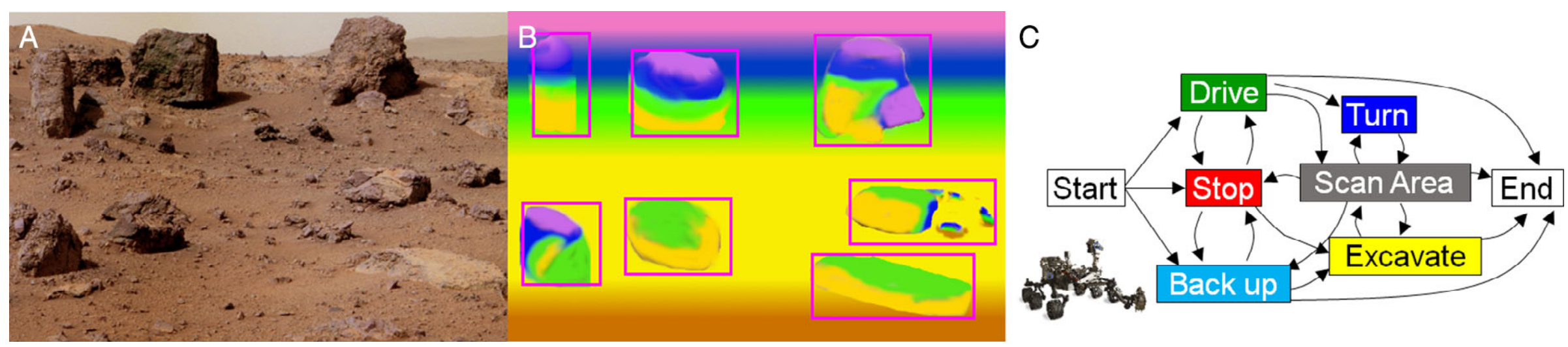

**Figure 2.** Dominant approach of geometry-based obstacle avoidance for robotic locomotion in complex environments. A) View from a Mars rover (https://mars.nasa.gov/mer/gallery/all/1/n/2695/1N367432321ESFBMLVP1961L0M1.JPG). B) Geometric map scanned. C) Driving modes to avoid obstacles (https://mars.nasa.gov/mars2020/spacecraft/rover/microphones/). Image credits: NASA/JPL-Caltech.

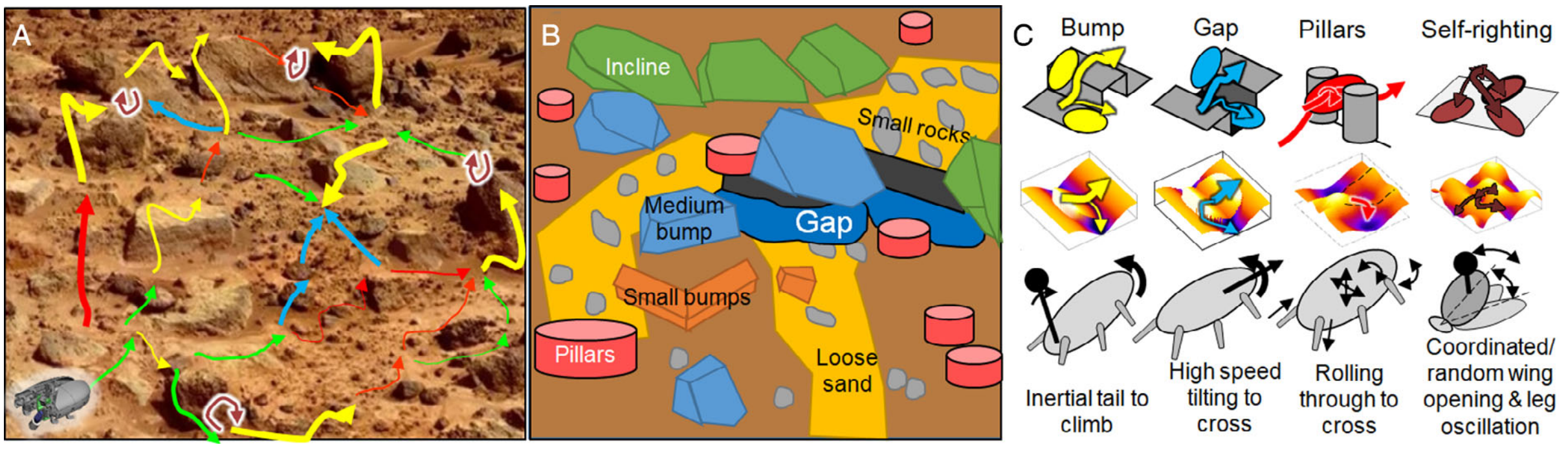

**Figure 3.** Envisioned locomotor transitions by bioinspired multilegged robots enabled by terradynamics of locomotor–terrain interaction with complex 3D terrain. A) Rocky Martian terrain of high scientific interest to be traversed by multilegged robots via locomotor transitions. Image credit: NASA/JPL-Caltech (https://mars.nasa.gov/MPF/parker/highres-stereo.html). B) Relevant large obstacle types abstracted. C) Discovered modulation of locomotor modes using actions predicted from physics models, which can be used to compose locomotor transitions in (A). Adapted under the terms of the CC-BY license.[119] Copyright 2021 The Authors. Published by the Royal Society.

Recent progress in principled understanding of legged and limbless locomotion in complex terrain has laid the scientific foundation for beginning to overcome this challenge. As demonstrated by a variety of animals with remarkable performance,[62–71] legged and limbless locomotion inherently offer advantages over wheeled locomotion in challenging terrain.[72–75] Early research on leg–ground interaction on relatively flat, rigid ground with small obstacles already elucidated the principles of generating and stabilizing single-mode locomotion like walking and running with high speed and efficiency.[76–84] These principles have enabled robust legged robot walking and running on such simple ground with performance approaching animals.[74,75] We have Boston Dynamics robots because we understand such interaction principles as well as we do.[84] Analogously, recent discoveries of the principles of legged walking and running on granular media[31,32,66,85–88,88–90] has laid the foundation for robust legged robot locomotion on sand. Similarly, fundamental understanding of locomotor–environment interaction for limbless locomotion using various gaits[68,91–96] has been instrumental in advancing the ability of snake-like robots to traverse flat surfaces, deformable terrain, and constrained environments.[68,72,73,97]

Here, we posit that biologically inspired legged and limbless robots are on the verge of providing more versatile alternative platforms for robustly traversing the aforementioned extraterrestrial terrain that are scientifically interesting yet difficult for wheeled rovers. Recent research in the Terradynamics Lab at Johns Hopkins University has further elucidated the principles of multilegged and limbless locomotion in complex 3D terrain with obstacles as large as the locomotor (animal or robot) themselves. We have systematically explored how animals such as insects and snakes and their robotic physical models physically interact with the terrain to generate effective locomotion (or lack thereof). Such principled understanding provides the scientific basis to apply or develop new robot design, control, planning, and machine-learning strategies to enable robust, autonomous traversal of cluttered large obstacles. An initial role of such robots could be a small scout to assess terrain traversability and collect samples for a larger rover.

For multilegged locomotion, we have discovered the principles of how to use and control physical interaction to generate locomotor transitions to traverse a diversity of large obstacles (e.g., Figure 3), ranging across those with height drop (gap)[98] and increase (bump)[99] and cluttered tall ones (pillars) with narrow spacing comparable to or smaller than the robot or animal,[100,101] as well as for self-righting after being flipped over,[102–107] a likely scenario during locomotion through large obstacles.[67,102,103] Across these diverse scenarios, stochastic yet stereotyped locomotor modes emerge (Figure 3C, top) as the self-propelled system is attracted to distinct basins of an underlying potential energy landscape resulting from locomotor–terrain interaction (Figure 3C, middle), which depends on the robot or animal's action. Thus, locomotor transitions can be generated by taking actions to destabilize the system to cross potential energy barriers on the landscape. Our physics modeling allowed interpretable, generalizable predictions of what actions a robot or animal can take to increase or decrease the probabilities of locomotor modes and transitions (Figure 3C, bottom) and how they depend on obstacle parameters (e.g., stiffness, friction, size, orientation), which have been validated experimentally. In addition, the potential energy barriers provide a proxy for the mechanical energetic cost of traversal, a key metric to optimize for space applications. For a more comprehensive review on our multilegged locomotion work, see Othayoth et al.[101]

We have recently created a proof-of-concept multimodal locomotion robot that integrates the various body and appendage designs and actions.[108] With human in the loop to trigger actions to switch between locomotor modes, we demonstrated that the robot is capable of traversing a diversity of obstacles representative of Martian rocks, with performance greatly exceeding what is possible by wheeled rovers, as well as self-righting after flipping over.[108] There are exciting opportunities to further develop physics-based planning and sensory feedback control strategies to enable autonomous multimodal locomotion. A plausible plan to achieve this is as follows. First, the robot will use our physics-based approach to form a task-level plan of how to take appropriate actions to compose a sequence of locomotor transitions to traverse various parts of the terrain between its landing site and scientifically interesting targets. This task-level plan can first be formed using terrain information that can be gathered beforehand (e.g., from satellite) and then continuously updated using more complete and refined information about terrain geometry and physical properties gathered during active traversal. Next, to robustly follow the planned trajectory using locomotor transitions, the robot must sense and react appropriately to unforeseen and uncharacterized variations of challenging terrain during active traversal, just like a self-driving car must sense and react to avoid unforeseen pedestrians or vehicles and uncharacterized roads. Besides vision cameras, force sensors must be added to the robot body and legs for sensing physical interaction with the terrain and determine the types and physical properties of the large obstacles encountered (see a proof of concept in Xuan et al.[109]). Discovering principles of bioinspired contact force sensory feedback control will facilitate this.[110] Reinforcement learning over a broad range of terrain variations can be added to physics-based models to enable precise sensory feedback control of actuation timing and magnitude required to execute the appropriate actions to make robust locomotor transitions. Finally, to assess advancements, we can measure traversal performance (traversal probability, mechanical energetic cost, speed) and compare it with that of the same robot fitted with wheels.

Recent deep learning approaches have begun to help multilegged robots learn to move over modest terrain by stabilizing the body in an upright posture.[111–113] However, the deep learning process is essentially an uninterpretable "black box" that cannot yield tractable insights and generalizable predictions of control policies for scenarios beyond those trained as our physics-based approach can. In addition, the learned policies only generate upright walking and running locomotion by stabilizing the body, limiting traversable obstacles to a small fraction of robot size.[111–113] By contrast, our approach can enable destabilizing locomotor transitions to traverse cluttered obstacles as large as the robot necessary for accessing scientifically interesting targets. This means our approach would achieve traversal of the same obstacle using a much smaller robot, saving payload for space launches; or it would achieve traversal of much larger obstacles using the same robot, significantly expanding the accessible terrain. Furthermore, a smaller robot is less susceptible to digging into loose sand due to reduced foot pressure from scaling.[89] Finally, although simulations are very useful for robotics,[114] experimental studies are indispensable for enabling robots to traverse terrain with complex interaction physics because pure simulations may not necessarily have the physics modeled correctly (i.e., simulation-to-reality gap[115]), whereas real robots enact, rather than model, the laws of physics[116–118]). Although pure learning approaches can, in principle, train the robot for any task in simulation by brute force, even in modest terrain, the real system's physics must still be modeled properly (e.g., how motor dynamics affects leg dynamics) to narrow the simulation-to-reality gap.[111–113] Thus, physics-based approaches must be added to make progress toward dynamic, destabilizing multimodal locomotion in real-world challenging terrain in which motion is highly sensitive to environmental perturbations and locomotor sensing and control imperfection and variation.[119]

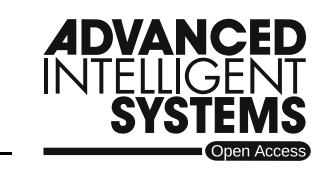

For limbless locomotion, our research has elucidated how to overcome stability challenges when traversing large obstacles that lack anchor points representative of rocks.[62,120,121] We have discovered that generalist snakes can control their long body to smoothly transition between lateral undulation which generates propulsion while maintaining static stability and vertical bending which bridges large height changes to adapt to large steps of a range of height and friction (**Figure 4**A, left, middle).[62] Thanks to its wide base of support, this simple adaptive gait has improved the traversal speed and probability of snake robots (Figure 4A, right) beyond previously achievable using simple follow-the-leader gaits and geometric motion planning.[120] For a more comprehensive review, see Fu and Li. More recently, it was discovered that vertical body bending can be used to generate propulsion to traverse large obstacles (Figure 4B, top) by both snakes and snake robots.[64,122] In addition, generalist snakes can smoothly transition between and adaptively coordinate lateral and vertical body bending to propel against multiple push points of a broad range of orientations to traverse uneven terrain filled with large obstacles with little transverse slip (Figure 4C).[123] The next step for limbless locomotion is to explore the principles of how to use sensory feedback control to modulate this adaptive 3D body bending coordination to generate overall directed propulsion. Recent animal studies of generalist snakes adapting lateral body bending to maintain contact and propulsion against large vertical obstacles on flat surfaces[124] suggested that this is a heavily sensory-modulated process, which likely involves proprioceptive and/or tactile sensing.[125] Toward understanding this, we have recently created a snake robot capable of 3D body bending with a compliant skin instrumented with distributed contact force sensors to detect physical interaction with the terrain during locomotion (Figure 4B, middle, bottom).[126]

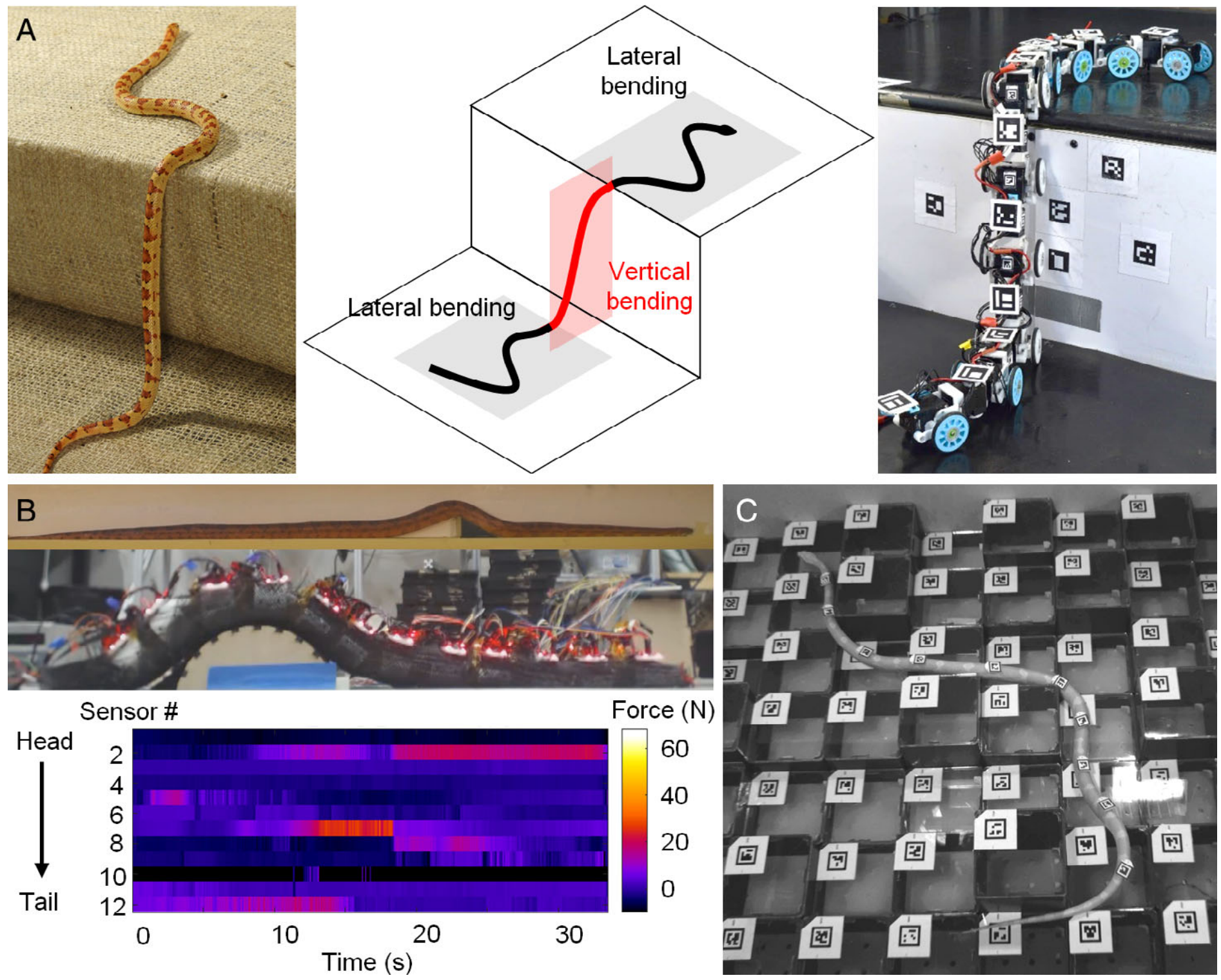

**Figure 4.** Progress in limbless locomotion traversing large obstacles. A) A simple gait transition template (middle) discovered from generalist snakes (left) offers high stability and enabled snake robots to traverse large smooth obstacles with high speed and probability (right). Adapted with permission.[121] Copyright 2020, Oxford University Press. B) Snake robot inspired from recent animal study (top) capable of using vertical body bending to traverse large obstacles (middle) as well as lateral bending, with the ability to sense contact forces (bottom). (Top) Adapted under the terms of the CC-BY license.[64] Copyright 2021 The Authors. Published by The Company of Biologists. C) Generalist snakes can combine vertical and lateral bending to traverse uneven terrain of large height variation.

To conclude, enabling new robotic locomotor capabilities to traverse challenging extreme terrain required for space scientific exploration is possible by continuing to broaden and deepen our understanding of the terradynamics of locomotor–terrain interaction in a diversity of complex terrain[31,67,101,119] (see discussion in Othayoth et al.). This progress can be facilitated by a systematic experimental physics approach to discover general principles of robot movement (i.e., robophysics[116,117]). Meanwhile, many efforts are also required for further engineering development and refinement of alternative platforms to become reliable under the harsh environmental conditions in space.[7] This advancement will also benefit other important mobile ground robot applications, such as search and rescue in earthquake rubble and environmental monitoring through forest floor debris and mountain boulders.

## Acknowledgements

C.L. thanks Eugene Lin for discussion and help with schematics. The authors thank two anonymous reviewers for suggestions. This work was supported by a Burroughs Wellcome Fund Career Award at the Scientific Interface and an Arnold & Mabel Beckman Foundation Beckman Young Investigator Award to C.L., and a Johns Hopkins University Space@Hopkins seed grant to C.L. and K.L.

## Conflict of Interest

The authors declare no conflict of interest.

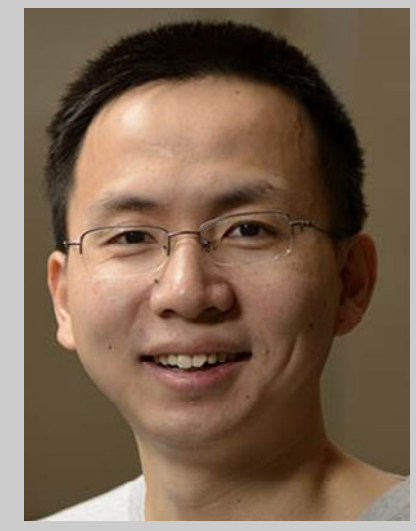

**Chen Li** is an assistant professor in the Department of Mechanical Engineering and Laboratory for Computational Sensing and Robotics at Johns Hopkins University. He earned B.S. and Ph.D. degrees in physics from Peking University and Georgia Tech, respectively, and did postdoctoral research in integrative biology and robotics at University of California, Berkeley. His research aims at creating the new field of terradynamics that describe the physics of locomotor–terrain interaction in complex terrain, analogous to aero- and hydrodynamics that describe fluid–structure interaction, and using terradynamics to understand animal locomotion and advance robot locomotion in the real world.

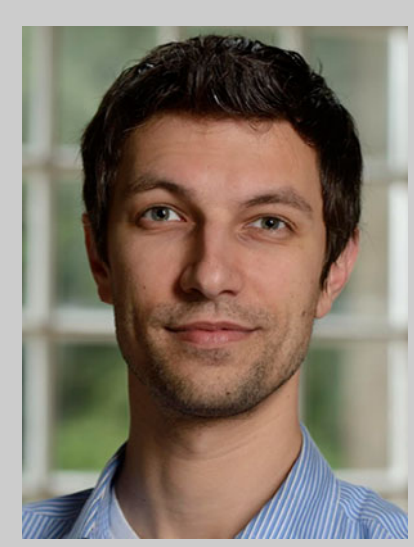

**Kevin Lewis** is an associate professor in the Department of Earth and Planetary Sciences at Johns Hopkins University. He earned B.S. in physics from Tufts and Ph.D. in planetary science from Caltech and did postdoctoral research at Princeton. His group studies the geology and geophysics of solid surface bodies in our solar system, and in particular the Moon and Mars. His research incorporates remote sensing and in situ data to interpret the geological evolution and modern surface conditions on these planetary bodies. He is a team member of the Mars Exploration Rover, Curiosity rover, and InSight lander missions to Mars.